\documentclass{article}

\usepackage[preprint]{neurips_2025}

\usepackage[utf8]{inputenc} 
\usepackage[T1]{fontenc}    
\usepackage{hyperref}       
\usepackage{url}            
\usepackage{booktabs}       
\usepackage{amsfonts}       
\usepackage{nicefrac}       
\usepackage{microtype}      
\usepackage{xcolor}         
\usepackage{wrapfig}
\usepackage{multirow}
\usepackage{amsmath}
\usepackage{mathtools}
\usepackage{graphicx}
\usepackage{subcaption}
\usepackage{algpseudocode}
\usepackage{bbm}
\usepackage{enumitem}

\usepackage{todonotes}

\usepackage[ruled,vlined,linesnumbered]{algorithm2e}
\newcommand{\tool}{\textsc{AutoLaw}}

\title{\textsc{AutoLaw}: Enhancing Legal Compliance in Large Language Models via Case Law Generation and Jury-Inspired Deliberation}

\author{
Tai D. Nguyen,~Long H. Pham,~Jun Sun \\
Singapore Management University, Singapore \\
\{dtnguyen.2019,hlpham,junsun\}@smu.edu.sg
}
\begin{document}

\maketitle

\begin{abstract}
The rapid advancement of domain-specific large language models (LLMs) in fields like law necessitates frameworks that account for nuanced regional legal distinctions, which are critical for ensuring compliance and trustworthiness. Existing legal evaluation benchmarks often lack adaptability and fail to address diverse local contexts, limiting their utility in dynamically evolving regulatory landscapes. To address these gaps, we propose \tool, a novel violation detection framework that combines adversarial data generation with a jury-inspired deliberation process to enhance legal compliance of LLMs. Unlike static approaches, \tool~dynamically synthesizes case law to reflect local regulations and employs a pool of LLM-based ``jurors'' to simulate judicial decision-making. Jurors are ranked and selected based on synthesized legal expertise, enabling a deliberation process that minimizes bias and improves detection accuracy. Evaluations across three benchmarks: Law-SG, Case-SG (legality), and Unfair-TOS (policy), demonstrate \tool's effectiveness: adversarial data generation improves LLM discrimination, while the jury-based voting strategy significantly boosts violation detection rates. Our results highlight the framework's ability to adaptively probe legal misalignments and deliver reliable, context-aware judgments, offering a scalable solution for evaluating and enhancing LLMs in legally sensitive applications.
\end{abstract}
\section{Introduction}

The rapidly improving capability of general-purpose large language models (LLMs) has accelerated the development of domain-specific LLMs for fields like biomedicine, finance, and law~\cite{gururangan2020don, cheng2023adapting}. While legal frameworks share common principles across jurisdictions, local nuances, such as Singapore's prohibition of ``eating on trains'' compared to its permissibility in the United States~\cite{eatingmrt}, demand meticulous handling to ensure LLMs provide accurate, regionally compliant responses. Failure to address these distinctions risks severe consequences, including regulatory violations, safety hazards, and erosion of user trust~\cite{wang2025comprehensive}. Yet, current methods for evaluating LLMs’ grasp of localized legal knowledge remain inadequate, raising urgent questions about their reliability in real-world applications.


A core challenge lies in the lack of adaptive, context-aware benchmarks. Existing datasets like LegalBench~\cite{guha2024legalbench} and LexGLUE~\cite{chalkidis2021lexglue} focus on broad legal tasks but inadequately represent jurisdictional diversity. Static benchmarks further struggle to keep pace with LLMs' rapid evolution, leaving emerging gaps in legal reasoning, such as subtle conflicts between regional policies, undetected~\cite{chang2024survey}. To address these limitations, we argue for the development of frameworks that dynamically evaluate models through automated, scenario-based probes, moving beyond fixed, labor-intensive evaluations.


We propose \tool, a novel framework that integrates {\bf adversarial legal data synthesis} with a {\bf jury-inspired deliberation mechanism} to enhance LLMs' legal compliance. Unlike traditional fine-tuning approaches, \tool~dynamically constructs a database of synthesized case law (i.e., historical judicial decisions tailored to local regulations) and assembles a ``jury'' of LLMs assigned specialized legal roles (e.g., Lawyer). These jurors are ranked by their expertise on synthesized case law and collaboratively deliberate on violation detection tasks, mimicking real-world judicial processes. By prioritizing adaptive reasoning over rigid parameter updates, \tool~minimizes biases while improving contextual accuracy.

\begin{wrapfigure}{r}{0.53\textwidth}
    \vspace{-0.8cm}
    \small
    \centering
    \begin{subfigure}[b]{0.25\textwidth}
        \includegraphics[width=\textwidth]{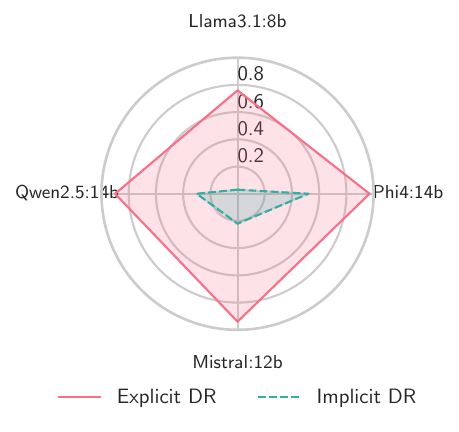}
        \caption{Data generation}
        \label{fig:show_case_data_generation}
    \end{subfigure}
    \begin{subfigure}[b]{0.25\textwidth}
        \includegraphics[width=0.9\textwidth]{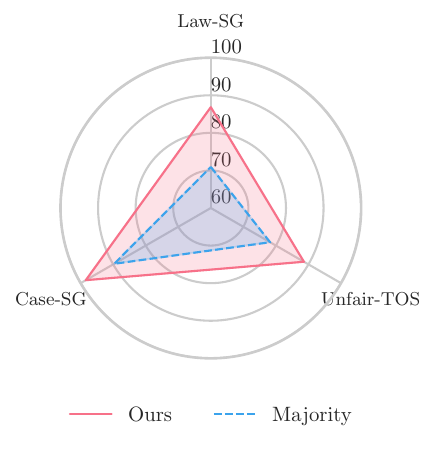}
        \caption{Violation detection}
        \label{fig:show_case_violation_detection}
    \end{subfigure}
    \caption{\small \tool~generates adversarial data and improves violation detection. Figure (a) shows detection rates (DR, the lower the better at discriminating LLMs) on explicit (trivial) and implicit (nuanced) data. Figure (b) presents violation detection rates (the higher the better performance) on three datasets.}
    \vspace{-0.3cm}
\end{wrapfigure}

We evaluate \tool~on three benchmarks spanning legality (Law-SG, Case-SG) and policy violation (Unfair-TOS~\cite{akash2024unfair}) scenarios. Our results demonstrate that: (1) {\bf adversarial data generation} effectively discriminates between LLMs' capabilities, exposing gaps in nuanced legal understanding (Figure~\ref{fig:show_case_data_generation}); (2) {\bf the jury deliberation strategy} increases violation detection accuracy over majority voting, while juror ranking ensures consistent reliability across diverse pools (Figure~\ref{fig:show_case_violation_detection}). These advancements position \tool~as a critical tool for developing legally robust LLMs, bridging the gap between static benchmarks and the dynamic demands of global compliance.

\section{Challenges}

To systematically evaluate techniques for enhancing LLMs’ understanding of region-specific legal frameworks, we adopt Singapore (a strict jurisdiction with distinct legal nuances) as a case study. Our investigation is structured around two pillars: (1) {\bf dataset generation}: we synthesize both explicit (trivial) and implicit (adversarial) legal scenarios to probe LLMs' grasp of Singaporean law; (2) {\bf violation detection enhancement}: we experiment with three complementary strategies: model editing, alignment, and retrieval-augmented generation (RAG) techniques.


\subsection{Data Generation}
\label{sect:data_generation}

\begin{figure}[htbp]
  \centering
  \begin{subfigure}[b]{0.55\textwidth}
    \centering
    \includegraphics[width=\textwidth]{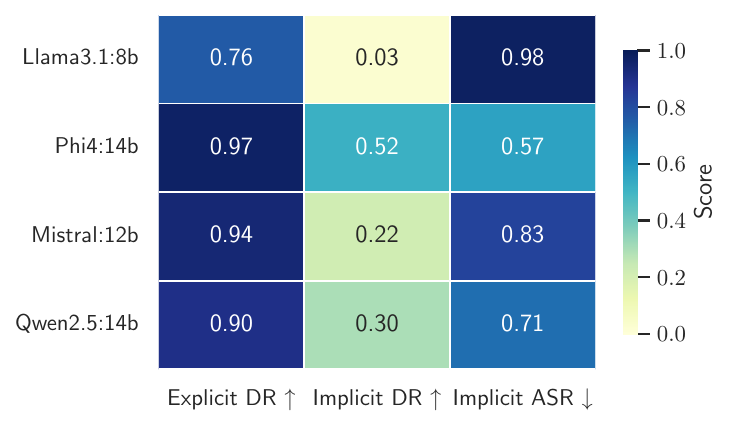}
    \caption{LLM performance on the violation detection task.}
    \label{fig:data_generation}
  \end{subfigure}
  \hfill
  \begin{subfigure}[b]{0.4\textwidth}
    \centering
    \includegraphics[width=\textwidth]{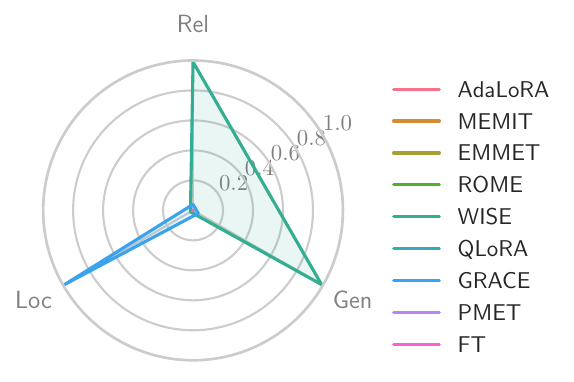}
    \caption{Editing performance on Llama3.1:8b}
    \label{fig:model_editing}
  \end{subfigure}
  \caption{LLM performance on the violation detection task and editing performance on Llama3.1:8b. In the left figure, explicit DR and implicit DR represent the detection rates on the direct data generation and adversarial data generation benchmarks, respectively. In in right figure, only WISE and GRACE show significant results across three metrics: rel, gen, and loc.}
  \label{fig:figure}
\end{figure}

\noindent{\bf Direct Data Generation Fails to Distinguish Model Capabilities.} To evaluate the effectiveness of LLMs in identifying region-specific legal violations, we first generate a baseline dataset using direct prompting. Twenty-five distinct regulatory violations are systematically selected from five Singaporean regulations (Table~\ref{tab:singapore_regulation_list}), and GPT-3.5 is prompted to generate 1,000 synthetic scenarios explicitly embedding these violations (see prompts in Table~\ref{tab:list_of_used_prompts}). Violation detection performance is evaluated using the detection rate (explicit DR), defined as the percentage of correctly identified violations, measured via a standardized prompting framework \cite{zheng2024ali}.


As illustrated in Figure~\ref{fig:data_generation}, four leading LLMs demonstrate remarkably high DRs (exceeding 0.76) on these synthetic scenarios. Notably, Phi4:14b achieves a near-ceiling DR of 0.97, indicating near-perfect identification of explicit violations. While these results confirm the proficiency of LLMs in recognizing explicit misconduct, the consistency in performance across models (with the variation $\Delta$DR < 0.21) reveals a critical limitation: {\it the dataset has limited capability to meaningfully differentiate model capabilities}. This highlights the inadequacy of the current approach and emphasizes the need for more sophisticated benchmarks that test models with subtle, contextually rich violations rather than relying on datasets generated through direct prompting.

\noindent{\bf Adversarial Testing Increases Discriminative Power.} Recent advances in adversarial testing~\cite{perez2022red, zheng2024ali, sun2024principle} have demonstrated its effectiveness in discriminating LLMs by probing their responses to long-tailed test inputs. Inspired by the adversarial framework ALI-Agent~\cite{zheng2024ali} and the principle-guided methodology of SELF-ALIGN~\cite{sun2024principle}, we employ an iterative refinement process to generate long-tailed scenarios, containing subtle but meaningful violations. These violations are designed to evade detection that prompts the target LLM with a specific violation detection prompt~\cite{zheng2024ali}.


We use GPT3.5 to refine 1000 initial scenarios, resulting in a dataset of 3,150 scenarios for evaluating four mainstream LLMs, including Llama3.1:8b, Phi4:14b, Mistral:12b, and Qwen2.5:14b. Figure~\ref{fig:data_generation} reveals significant model disparities through two metrics: detection rate (implicit DR) and attack success rate (ASR). The ASR measures how often GPT3.5 can successfully refine a scenario. The observed DR ranges widely, from 0.03 to 0.52, showing substantial variation across LLMs. The ASR ranges from 0.57 to 0.98, with Llama3.1:8b being most vulnerable to adversarial testing. In contrast, the 14B-parameter models, including Qwen2.5:14b and Phi4:14b prove more resistant, with ASRs of 0.57 and 0.71, respectively. Larger models demonstrate better attack resistance and higher detection rates. This wide discrepancy highlights the efficacy of the adversarial dataset in discriminating between model capabilities.

\subsection{Violation Detection}
\label{sect:violation_detection}
To improve an LLM's capability of understanding region-specific law, we adopt and evaluate the following popular model customization techniques in our study.

\begin{figure}[t]
    \centering
    \includegraphics[width=0.7\textwidth]{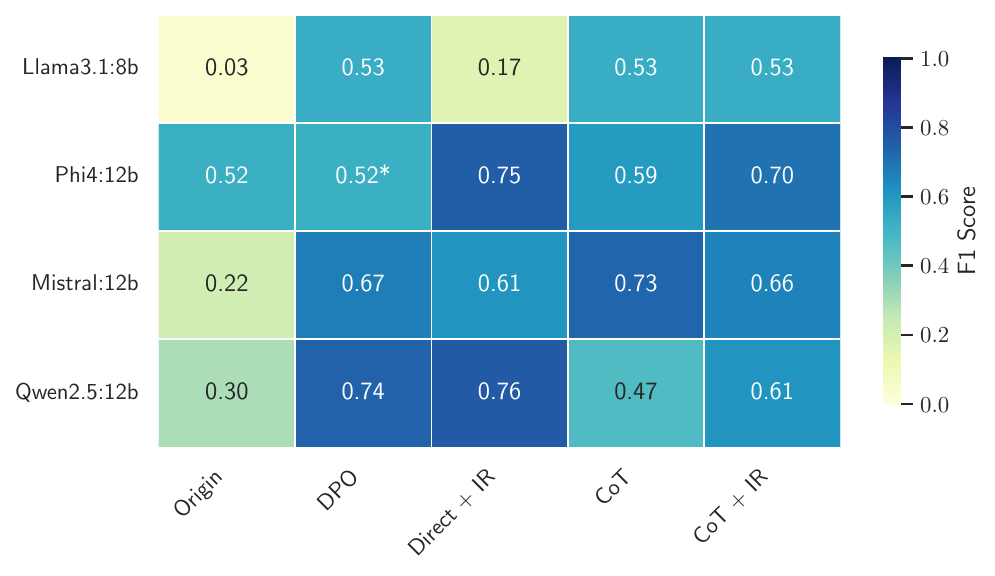}
    \caption{Performance comparison of LLMs with DPO and RAG enhancements. The \textbf{origin} represents baseline LLM performance without DPO or RAG.}
    \label{fig:dpo_and_rag}
\end{figure}

\noindent{\bf Model Editing Harms General Abilities of LLMs.} Model editing has emerged as an efficient technique for dynamically updating knowledge in LLMs. To assess its potential for improving law violation detection performance, we systematically evaluate ten model editing methods on Llama3.1 using 3,150 successful refinement scenarios from the above-mentioned studies. These scenarios are partitioned into training (80\%) and testing (20\%) subsets, with performance rigorously evaluated across three established metrics~\cite{Meng2022LocatingAE}: {\it reliability} (the ability to recall updated knowledge), {\it generality} (the ability to generalize updated knowledge), and {\it locality} (preservation of unrelated information). For the locality evaluation, we construct a complementary dataset by sampling an equivalent number of test instances from the DROP~\cite{dua2019drop} benchmark, ensuring a balanced comparison of the model's ability to retain unmodified reasoning skills (see Sect.~\ref{sect:model_editing_details} for editing details).


As shown in Figure~\ref{fig:model_editing}, the experimental results highlight a stark contrast between the editing method named WISE~\cite{wang2024wise} and other model editing methods. While WISE achieves high reliability and generality, its significantly lower locality score suggests overfitting, likely due to its tendency to respond ``Yes'' to all input scenarios uniformly. This pattern indicates that WISE prioritizes correctness in edited cases at the expense of preserving unrelated knowledge. In contrast, other methods severely degrade model performance, rendering outputs unreliable or unusable across all evaluated metrics. The results indicate that model editing impairs the general capabilities of LLMs, rendering them unsuitable for the violation detection task.

\noindent{\bf Alignment Methods Effectively Detect Violations.} Direct Preference Optimization (DPO)~\cite{rafailov2023direct}, have garnered significant interest due to their scalability and ease of use. To evaluate whether DPO can improve LLMs' ability to detect violations, we assess DPO across four mainstream LLMs using 3150 refinement scenarios used in the above study. These scenarios are partitioned into training (80\%) and testing (20\%) subsets, with model performance measured using $F_1$ (see Sect.~\ref{sect:alignment_details} for alignment details). 
Figure~\ref{fig:dpo_and_rag} reveals a notable enhancement in performance, with $F_1$ scores rising across all models~\footnote{$0.52^*$ indicates a failure to perform DPO on Phi4 due to architectural differences between the Unsloth~\cite{unsloth} Phi4 and the original release.}
. Among them, Qwen2.5:14b attains the highest $F_1$ score after applying DPO, whereas Phi4:12b excels prior to DPO. The most significant leap, however, is observed in Llama3.1:8b, which records the greatest improvement with a delta of 0.5. These findings underscore the effectiveness of DPO in bolstering the violation detection capabilities of the pretrained models.

\noindent{\bf Retrieval-Augmented Generation Yields Impressive Gains.} Another approach to improve violation detection is Retrieval-Augmented Generation (RAG), a method that enhances the reasoning ability of LLMs by supplementing them with external knowledge. To evaluate RAG's effectiveness, we utilize the same dataset from the prior experiment and compare $F_1$ scores across four configurations: violation detection prompt~\cite{zheng2024ali} (Direct), Zero-Shot-Cot (CoT) without information retrieval (IR), and CoT with IR (see Sect.~\ref{sect:rag_details} for RAG details).

Figure~\ref{fig:dpo_and_rag} shows a notable performance improvement when using Zero-Shot-CoT compared to the baseline in all cases. Surprisingly, CoT with IR in Mistral:12b yields a lower $F_1$ score than CoT. Further analysis indicates that the provided context is misleading, causing the LLMs to misinterpret the question. For example, when a scenario involving MRT misconduct is paired with context about liquor-related violations, the LLM may incorrectly conclude that no violation has occurred, correctly identifying the absence of liquor misconduct, but failing to recognize the actual MRT-related offense. The results confirm that RAG enhances violation detection, but its effectiveness hinges on the quality of the provided context.
\section{Method of \tool}
\label{sect:method}
\begin{figure}[]
  \centering
  \begin{subfigure}[b]{0.27\textwidth}
    \centering
    \includegraphics[width=\textwidth]{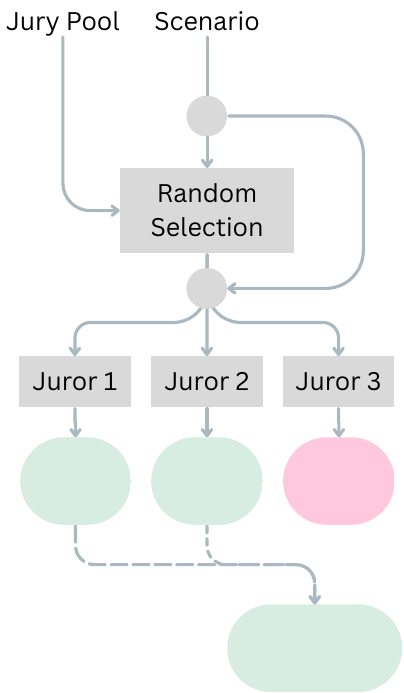}
    \caption{Majority Vote}
    \label{fig:image2}
  \end{subfigure}
  \hfill
  \begin{subfigure}[b]{0.62\textwidth}
    \centering
    \includegraphics[width=\textwidth]{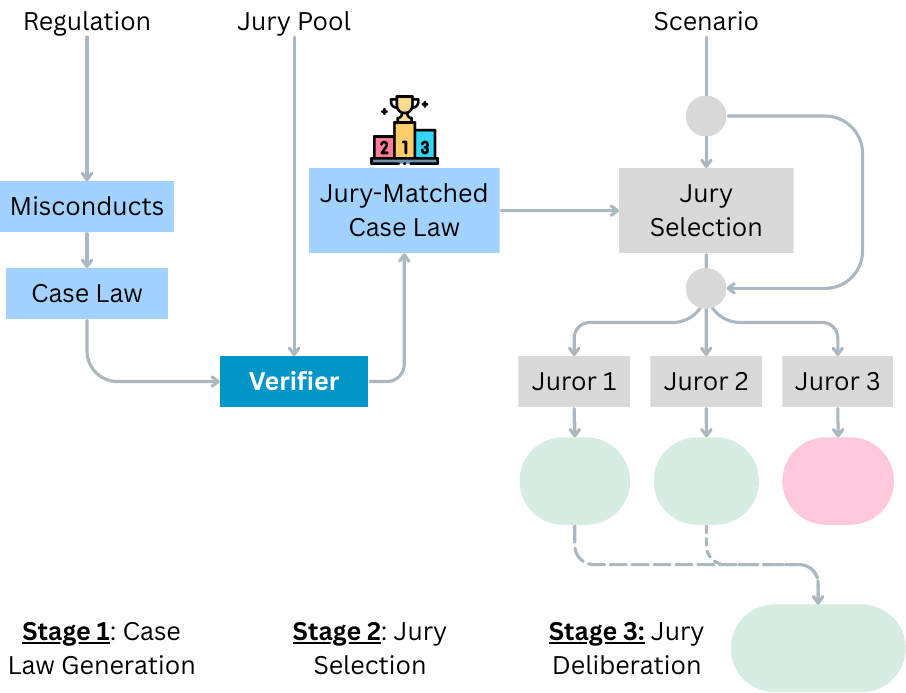}
    \caption{\tool}
    \label{fig:auto_law}
  \end{subfigure}
  \caption{An overview of the existing majority vote (a) and the proposed \tool~(b). The majority vote approach randomly selects jurors from a juror pool and adopts the answer with the most votes. In contrast, \tool~selects the top $k$ jurors from the pool for each scenario, using generated scenarios as demonstrations to improve detection performance. Additionally, the \underline{Stage 1} and \underline{Stage 2} are pre-computed offline to minimize the runtime overhead during ranking.}
  \label{fig:workflow}
\end{figure}

Based on the observations and considerations identified in Sect 2, we propose a law-abiding system \tool~to mitigate bias amplification and reliability risks inherent in standalone alignment methods and RAG systems.  Our framework automatically strengthens the violation detection ability of LLMs for local regulations through three optimized stages (Figure~\ref{fig:auto_law}): {\bf (stage 1) case law generation}: extracts actionable misconducts from regulations and generates violated scenarios through an adversarial testing process, {\bf (stage 2) jury selection}: ranks jury pool members for each case law according to the correctness of their results, {\bf (stage 3) jury deliberation}: aggregates votes from a selected jury to determine the verdict. An example illustrating the calculation at each stage is provided in Table~\ref{tab:step_calculation}.

\begin{table}[t]
    \centering
    \small
    \begin{tabular}{p{0.8cm}p{2.5cm}p{9.3cm}}
    \toprule
    {\bf Stage} & {\bf Name} & {\bf Description} \\
    \midrule
    \multirow{3}{*}{Stage 1} & Regulation -- $r$ & Singapore Rapid Transit Systems. \\
    & Misconduct -- $m$ & \colorbox[HTML]{a2d2ff}{Consuming food or drinks except in designated areas.} \\
    & Case law -- $\text{ALI}(m)$ & \textcolor[HTML]{a2d2ff}{A commuter {\it \underline{sneaks a quick bite of a sandwich}} while riding the train, trying to {\it \underline{discreetly finish their meal}} before reaching their destination.} \\
    \midrule
    \multirow{3}{*}{Stage 2} & Jury pool - $P$ & {\it (Judge, Llama3.1), (Prosecutor, Llama3.1), \textbf{(Judge, Qwen2.5)}, \textbf{(Lawyer, Qwen2.5)}, \textbf{(Prosecutor, Phi4)}, (Judge, Mistral)}. \\
    & Verifier - $\mathcal{V}(m,r)$ & \textcolor[HTML]{0096c7}{$(0.2, 0.5, 0.3, \textbf{1.0}, \textbf{0.9}, \textbf{0.8}, 0.7)$}.\\
    & Jury - $\mathcal{J}(x)$ & \colorbox[HTML]{d9d9d9}{\it \textbf{(Judge, Qwen2.5)}, \textbf{(Lawyer,
Qwen2.5)}, \textbf{(Prosecutor, Phi4)}}. \\
    \midrule
    \multirow{3}{*}{Stage 3} & Input -- $x$ & Sarah quickly unwrapped her sandwich and took a bite on the train due to a medical condition that required her to eat immediately. 
    \\
    & Votes & {\bf \colorbox[HTML]{d7ede2}{Yes}, \colorbox[HTML]{d7ede2}{Yes}, \colorbox[HTML]{ffc8dd}{No}}. \\
    & Output -- $y$ & \colorbox[HTML]{d7ede2}{{\bf Yes}}. \\
    \bottomrule
    \end{tabular}
    \vspace{0.3cm}
    \caption{Examples of different components in~\tool. The highlighted colors match with the colors in the Fig.~\ref{fig:auto_law}. To handle the input $x$, three jurors are selected from a jury pool of six according to the ranking scores produced by the $\mathcal{V}$. The aggregated output is \colorbox[HTML]{d7ede2}{{\bf Yes}}.}
    \label{tab:step_calculation}
\end{table}

\begin{figure}[t]
\begin{minipage}{0.5\textwidth}
\begin{algorithm}[H]
\SetAlgoLined
\SetKwInOut{Input}{Input}
\SetKwInOut{Output}{Output}
\Input{Regulation $R$ \\ Jury pool $P$}
let $\mathcal{C}^* \gets \emptyset$ \;
\ForEach{$r \in R$}{
    \ForEach{$m \gets \text{LLM}(p \oplus r)$}{
        \ForEach{$x \in \text{ALI}(m)$}{
            $\mathcal{C}^* \gets \mathcal{C}^* \cup \{(r, m, x, \mathcal{J}(x))\}$ \;
        }
    }
}
\Output{Jury-matched case law $\mathcal{C}^*$}
\caption{Jury Selection + CL Generation}
\label{algo:stage_1_2}
\end{algorithm}
\end{minipage}
\hfill
\begin{minipage}{0.46\textwidth}
\begin{algorithm}[H]
\SetAlgoLined
\SetKwInOut{Input}{Input}
\SetKwInOut{Output}{Output}
\Input{Jury-matched case law $\mathcal{C}^*$ \\ Threshold $\theta$ \\ Input scenario $x$}
\text{let } $y \gets false$ \;
\text{let } $votes \gets 0$ \;
\text{let } $\hat{x} \gets \arg\max_{\hat{x} \in \mathcal{C}^*} \text{sim}(x, \hat{x})$ \;
\ForEach{$J \in \mathcal{J}(\hat{x})$}{
    $votes \gets votes + \mathcal{F} \circ J(p \oplus \hat{x} \oplus x)$ \;
}
\text{let } $y \gets votes > \theta * |\mathcal{J}(\hat{x})|$ \;
\Output{Verdict $y$}
\caption{Jury Deliberation}
\label{algo:stage_3}
\end{algorithm}
\end{minipage}
\end{figure}

\subsection{Case Law Generation}
A case law dataset is generated from a set of regulations $R$ through the following process. First, each regulation $r \in R$ is analyzed to extract a list of potential misconducts. Next, each misconduct $m$ undergoes a refinement loop using the ALI method~\cite{zheng2024ali} to produce long-tailed scenarios $x$. These long-tailed cases, along with their corresponding misconducts and regulations, are compiled into the dataset:
\[
\mathcal{C} = \{(x, m, r) \mid x \in \text{ALI}(m) \land m \in \text{LLM}(p \oplus r) \land r \in R\}
\]
where $\text{LLM}(p \oplus r)$ denotes the extraction of misconducts using a language model, and $\text{ALI}(m)$ represents the generation of edge case scenarios from misconduct. For example, from the misconduct {\it ``Bringing dangerous or flammable substances.''} of a regulation {\it``Singapore Rapid Transit Systems''}, we generate a scenario such as {\it ``Sarah wanted to surprise her friends with fireworks at a gathering. As she boarded the MRT, the fireworks accidentally ignited, causing a brief moment of surprise among passengers.''}

\subsection{Jury Selection}

A jury pool $P$ is formed by integrating existing LLMs (both pre-trained and fine-tuned) and assigning them specific roles. For instance, a pre-trained Qwen2.5 may serve as a judge, while a fine-tuned Llama3.1 could act as a lawyer. All potential jurors, denoted as $J$, undergo a rigorous evaluation to assess their legal knowledge and suitability. The evaluation process is formalized as follows:
\[
\mathcal{J}(x) = \arg\max_{\substack{S \subseteq P \\ |S| = k}} \sum_{J \in S} \mathcal{V}(m, r) \circ J(p \oplus x), \quad \text{where } (x, m, r) \in \mathcal{C},\ k \in \mathbb{N},\ k > 0
\]
where $x$ represents a scenario that belongs to a case law dataset $\mathcal{C}$. The top $k$ jurors with the highest scores are selected for $\mathcal{J}(x)$. The verifier $\mathcal{V}(m,r)$ is an evaluator function that evaluates the juror's response, assigning a correctness score from $0.0$ to $1.0$. Powered by advanced language models (e.g., GPT4.1), the verifier is capable of performing sophisticated legal analysis. By analyzing the alleged misconduct and the relevant regulation, the verifier accurately assesses the correctness of a given answer.



\subsection{Jury Deliberation}

Each jury member independently evaluates the provided scenario $x_i$ and submits their answer. These answers are aggregated to determine the outcome. If more than half of the jury members vote ``Yes'', indicating a violation, the outcome is ``Yes''. Otherwise, the outcome is ``No''. The voting process is formalized as:
\[
\begin{aligned}
&\text{Let } \hat{x} = \arg\max_{\hat{x} \in \mathcal{C}} \text{sim}(x, \hat{x}). \\
&\text{Then } y =
\begin{cases}
1 & \text{if } \frac{1}{|\mathcal{J}(\hat{x})|} \sum\limits_{J \in \mathcal{J}(\hat{x})} \mathcal{F} \circ J(p \oplus \hat{x} \oplus x) > \theta \\
0 & \text{otherwise}
\end{cases}
\quad \text{for threshold } \theta = 0.5.
\end{aligned}
\]

where $\mathcal{F}$ is an evaluator function that returns 1 for violations and 0 for compliant cases. The constant $\theta = 0.5$ enforces a majority voting mechanism. The function {\it sim} denotes the semantic similarity function, which maps a pair of scenarios to a score between 0.0 and 1.0. The most similar scenario $\hat{x}$ is a demonstration. The detailed algorithm of \tool~is shown in Algorithm~\ref{algo:stage_1_2} and Algorithm~\ref{algo:stage_3}, where Jury-matched case law is case law that contains the ranked jury pool.

\section{Experiment}
In this section, we conduct multiple experiments, aiming  to answer the following research questions:
\begin{itemize}
    \item \textbf{RQ1}: How effectively does \tool~detect violated scenarios?
    \item \textbf{RQ2}: How well does \tool~detect violations in real-world scenarios?
    \item \textbf{RQ3}: How do the components of \tool~influence its overall effectiveness?
\end{itemize}
\noindent{\bf Datasets.} To verify \tool's effectiveness as a violation detection framework. We conduct experiments on three datasets covering two key violation types: legality ({\it Law-SG and Case-SG}) and policy ({\it Unfair-TOS}). Law-SG was systematically generated using ALI~\cite{zheng2024ali} by targeting four LLMs, while Case-SG was curated from Singapore newspapers. Unfair-TOS is a subset of LexGLUE~\cite{chalkidis2021lexglue} (see Appendix \ref{sect:datasets} for detailed descriptions of the datasets).

\noindent{\bf Evaluation Metrics.} To evaluate the violation detection capability of LLMs, we adopt a standard metric {\it detection rate}~\cite{zheng2024ali} (DR), which ranges from 0.00 to 100. Detection rate is the percentage of times that the LLM correctly identifies the violated scenarios. A higher detection rate indicates a better identification ability. We apply this metric to all three datasets.

\noindent{\bf Baselines.} We compare \tool~with majority vote (MV) across three jury pools, three legal roles, and seven LLMs. Each jury pool is randomly formed by selecting combinations of LLMs and legal roles. For every dataset, we sample a new set of jury pools. Vote-x (where x = $1,3,5$) denotes a voting setup with x jurors.

\noindent{\bf Jury Pool Setup.} We select four mainstream LLMs along with their corresponding fine-tuned variants, each assigned roles is one of three legal domain: {\it Judge, Lawyer, and Prosecutor}. This results in a pool of 21 potential jurors. To form a x-member jury, we first randomly sample six jurors from this pool, simulating the summoning of potential jurors. These six jurors then undergo a jury selection process to finalize the x-member jury. The details of each jury pool are presented in Table~\ref{tab:jury_pools}.

\begin{table}[t]
\begin{subtable}{0.49\textwidth}
\resizebox{\textwidth}{!}{%
    \centering
    \begin{tabular}{cccccccc}
    \toprule
    {\bf Settings}
    & \multicolumn{6}{c}{{\bf Law-SG} -- {\it detection rate} $\uparrow$} \\
    {\it Zero-Shot-CoT} & \multicolumn{2}{c}{\it $P_1$} & \multicolumn{2}{c}{\it $P_2$} & \multicolumn{2}{c}{\it $P_3$} \\
    \midrule
    & MV & Ours & MV & Ours & MV & Ours\\
    \cmidrule(l){2-3}
    \cmidrule(l){4-5}
    \cmidrule(l){6-7}
    {\it Vote-1} & 63.35 & {\bf 85.62} & 73.14 & {\bf 87.52} & 63.19 & {\bf 83.25} \\
    {\it Vote-3} & 66.19 & {\bf 86.73} & 76.94  & {\bf 88.31} & 64.93 & {\bf 84.52} \\
    {\it Vote-5} & 67.46 & {\bf 85.31} & 78.20  & {\bf 90.36} & 67.14 & {\bf 84.83} \\
    \midrule
    J1 & \multicolumn{2}{c}{\underline{71.88}}&  \multicolumn{2}{c}{70.14}& \multicolumn{2}{c}{53.08} \\
    J2 & \multicolumn{2}{c}{\bf 78.52}& \multicolumn{2}{c}{\underline{78.52}}&\multicolumn{2}{c}{\underline{66.98}} \\
    J3 & \multicolumn{2}{c}{53.40}&\multicolumn{2}{c}{\bf 80.09}&\multicolumn{2}{c}{\bf 72.20} \\
    J4 & \multicolumn{2}{c}{66.67}&\multicolumn{2}{c}{66.51}&\multicolumn{2}{c}{50.87} \\
    J5 & \multicolumn{2}{c}{51.66}&\multicolumn{2}{c}{74.25}&\multicolumn{2}{c}{66.67} \\
    J6 & \multicolumn{2}{c}{67.14}&\multicolumn{2}{c}{67.61}&\multicolumn{2}{c}{\bf 72.20} \\
    \bottomrule
    \end{tabular}
}
\subcaption{Legality}
\label{tab:legality}
\end{subtable}
\hfill
\begin{subtable}{0.49\textwidth}
\resizebox{\textwidth}{!}{%
    \centering
    \begin{tabular}{cccccccc}
    \toprule
    {\bf Settings}
    & \multicolumn{6}{c}{{\bf Unfair-TOS}~\cite{akash2024unfair} -- {\it detection rate} $\uparrow$} \\
    {\it Zero-Shot-CoT} & \multicolumn{2}{c}{\it $P_4$} & \multicolumn{2}{c}{\it $P_5$} & \multicolumn{2}{c}{\it $P_6$} \\
    \midrule
    & MV & Ours & MV & Ours & MV & Ours\\
    \cmidrule(l){2-3}
    \cmidrule(l){4-5}
    \cmidrule(l){6-7}
    {\it Vote-1} & 64.94 & {\bf 87.01} & 69.26 & {\bf 83.98} & 74.89 & {\bf 87.01} \\
    {\it Vote-3} & 68.83 & {\bf 86.58} & 77.92 & {\bf 88.74} & 79.65 & {\bf 86.58} \\
    {\it Vote-5} & 73.59 & {\bf 88.31} & 80.09 & {\bf 88.74} & 80.95 & {\bf 88.74} \\
    \midrule
    J1 & \multicolumn{2}{c}{70.13} & \multicolumn{2}{c}{66.67} & \multicolumn{2}{c}{\underline{78.35}} \\
    J2 & \multicolumn{2}{c}{\underline{78.35}} & \multicolumn{2}{c}{\bf 90.48} & \multicolumn{2}{c}{61.47} \\
    J3 & \multicolumn{2}{c}{\bf 82.68} & \multicolumn{2}{c}{72.29} & \multicolumn{2}{c}{70.13} \\
    J4 & \multicolumn{2}{c}{69.70} & \multicolumn{2}{c}{79.22} & \multicolumn{2}{c}{74.46} \\
    J5 & \multicolumn{2}{c}{31.60} & \multicolumn{2}{c}{\underline{84.42}} & \multicolumn{2}{c}{71.86} \\
    J6 & \multicolumn{2}{c}{61.04} & \multicolumn{2}{c}{33.77} & \multicolumn{2}{c}{\bf 92.21} \\
    \bottomrule
    \end{tabular}
}
\caption{Policy}
\label{tab:policy}
\end{subtable}
\caption{The performance comparison on Legality and Policy.}
\vspace{-0.5cm}
\end{table}



\subsection{Performance Comparison (RQ1)}
\label{sect:rq1}
\paragraph{Results.} {
Tables~\ref{tab:legality}, \ref{tab:policy} show the detection rate of six jury pools ($P_1$-$P_6$) on two datasets (Law-SG and Unfair-TOS) across all evaluation settings. Individual detection rates for jurors J1–J6 within each pool are also shown. {\bf Bold} numbers indicate the highest detection rates, while \underline{underlined} numbers represent the second-highest. From the results, we have the following key observations:
\begin{itemize}[leftmargin=*, nosep]
    \item {\bf \tool~detects more violated cases compared to majority vote in both legal and policy datasets.} The results reveal that \tool~achieves the highest detection rates across all cases. The performance gaps between \tool~and the majority vote baselines are substantial, ranging from 11.37 ({\it Vote-3}, $P_2$) to 22.27 ({\it Vote-1}, $P_1$). The empirical results demonstrate \tool's effectiveness in detecting violations, highlighting the importance of collaboration among LLMs to leverage each model's strengths.
    \item {\bf \tool~is reliable.} By comparing voting results across three pool, particularly Vote-5, we observe that \tool~reduces detection rate variance. For example, the standard deviation for majority vote is 6.30 and 4.02, while \tool~achieves lower values of 3.06 and 0.25 in Tables~\ref{tab:legality} and \ref{tab:policy}. These results demonstrate \tool's superior consistency across pools.
    \item {\bf Larger jury pools consistently lead to improved detection performance.} The results exhibit a common performance pattern: Vote-1 < Vote-3 < Vote-5, observed in both \tool~and majority vote. This trend is more pronounced in the majority vote, where jurors are randomly selected. In contrast, \tool's performance is influenced by additional factors, such as juror roles and relevant case law. These findings highlight the advantage of larger jury pools in improving detection accuracy across diverse jury settings.
    \item {\bf Majority vote in heterogeneous LLMs results into average performance.} The detection rates vary across models J1–J6, confirming their heterogeneity. Unlike self-consistency~\cite{wang2022self}, which improves both reliability and accuracy, majority voting in our setting yields only average results. This behavior is expected, as combining predictions from both strong and weak models typically leads to diluted performance. These findings highlight the necessity of our approach, which aims to enhance voting mechanisms in heterogeneous LLM ensembles.
\end{itemize}
}

\subsection{Performance Comparison (RQ2)}
\begin{table}[t]
    \centering
    \small
    \begin{tabular}{cccccccc}
    \toprule
    {\bf Evaluation Settings}
    & \multicolumn{6}{c}{{\bf Case-SG} -- {\it detection rate} $\uparrow$} \\
    {\it Zero-Shot-CoT} & \multicolumn{2}{c}{\it $P_7$} & \multicolumn{2}{c}{\it $P_8$} & \multicolumn{2}{c}{\it $P_9$} \\
    \midrule
    & MV & Ours & MV & Ours & MV & Ours\\
    \cmidrule(l){2-3}
    \cmidrule(l){4-5}
    \cmidrule(l){6-7}
    {\it Vote-1} & 88.10 &{\bf 100.0}&88.10&{\bf 100.0}&85.71 & {\bf 97.62} \\
    {\it Vote-3} & 92.86 &{\bf 97.62}&85.71&{\bf 97.62}&90.48 & {\bf 100.0} \\
    {\it Vote-5} & 92.86 &{\bf 97.62}&85.71&{\bf 97.62}&90.48 & {\bf 100.0} \\
    \midrule
    J1 & \multicolumn{2}{c}{88.10} & \multicolumn{2}{c}{\bf 90.48} & \multicolumn{2}{c}{\underline{90.48}} \\
    J2 & \multicolumn{2}{c}{\bf 95.24} & \multicolumn{2}{c}{\bf 90.48} & \multicolumn{2}{c}{88.10} \\
    J3 & \multicolumn{2}{c}{85.71} & \multicolumn{2}{c}{\underline{85.71}} & \multicolumn{2}{c}{\bf 92.86} \\
    J4 & \multicolumn{2}{c}{\underline{92.86}} & \multicolumn{2}{c}{\bf 90.48} & \multicolumn{2}{c}{\underline{90.48}} \\
    J5 & \multicolumn{2}{c}{90.48} & \multicolumn{2}{c}{80.95} & \multicolumn{2}{c}{83.33} \\
    J6 & \multicolumn{2}{c}{90.48} & \multicolumn{2}{c}{83.33} & \multicolumn{2}{c}{80.95} \\
    \bottomrule
    \end{tabular}
    \vspace{0.3cm}
    \caption{The performance comparison on real-world case law in Singapore}
    \label{tab:case_sg}
\end{table}
\noindent{\bf Motivations.} As shown in Section~\ref{sect:rq1}, \tool~has proven highly effective in uncovering violations deeply embedded within complex, long-tailed scenarios, showcasing its robust detection capabilities across diverse and challenging contexts. Motivated by this success, RQ2 seeks to explore \tool~'s applicability in real-world settings. By evaluating \tool~'s performance in practical scenarios, we aim to assess its potential to assist legal professionals in identifying violations efficiently.

\noindent{\bf Results.} Table~\ref{tab:case_sg} shows the detection rate of \tool~on the Case-SG dataset. Individual detection rates for jurors J1–J6 within each pool are also shown. {\bf Bold} numbers indicate the highest detection rates, while \underline{underlined} numbers represent the second-highest. We observe that individual jurors (J1-J6) achieve consistently high detection rates, with minimum scores of 88.10, 80.95, and 80.95 in pools $P_7$, $P_8$, and $P_9$ respectively. This strong performance enables \tool~to achieve a perfect score of 100 in four voting settings, and 97.62 in the remaining ones. It consistently outperforms the majority vote baseline. A closer investigation reveals that real-world case law is typically easy to detect due to two main reasons. First, the violation is often explicit, as in {\it a man was seen vaping openly on an MRT train while leaning against the doors''}. Second, such cases are often accompanied by clear enforcement signals, such as {\it was issued a fine by enforcement officers''}. These results suggest that \tool~is highly effective in identifying legal violations in realistic, regulation-grounded scenarios.

\subsection{Ablation Studies (RQ3)}
\begin{figure}[t]
    \centering
    \includegraphics[width=0.9\linewidth]{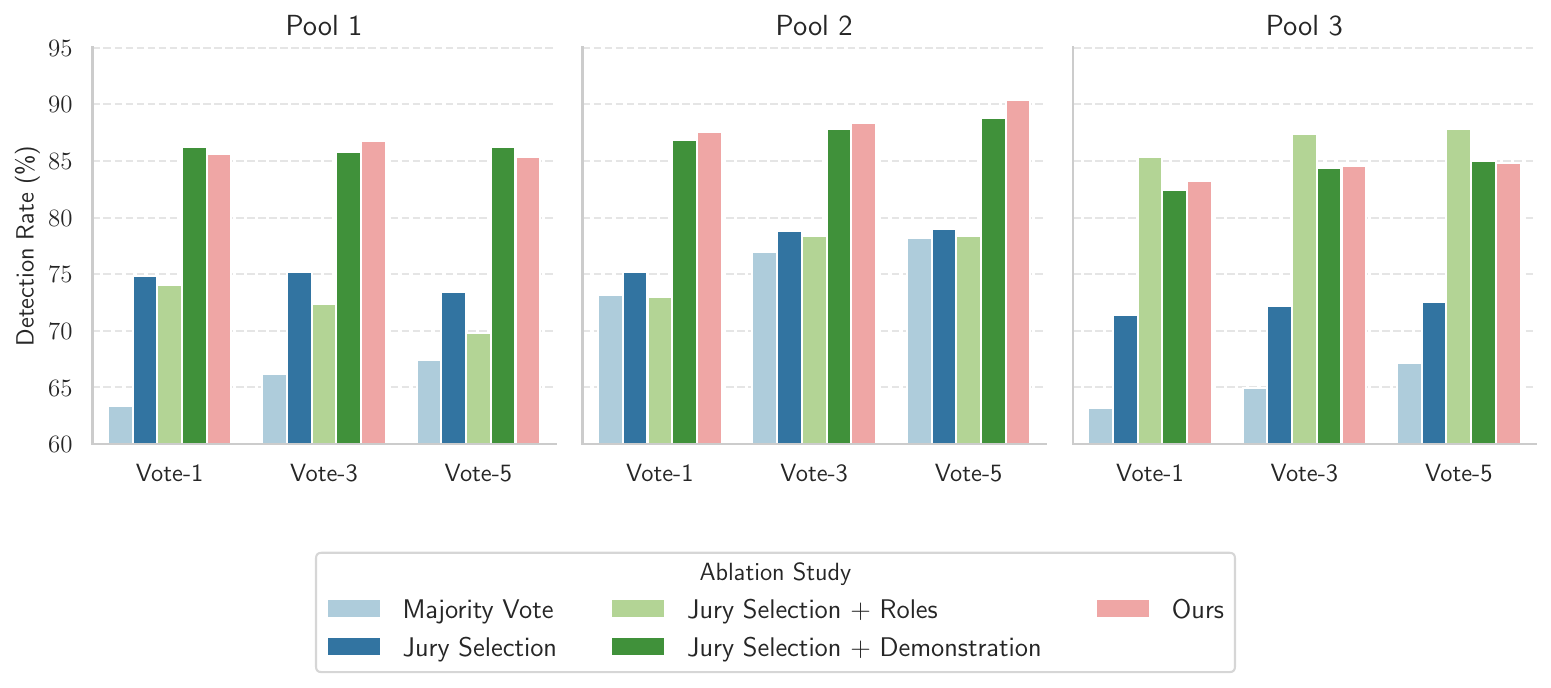}
    \caption{The impact of each component (e.g., jury selection)  on Law-SG dataset.}
    \label{fig:ablation_study}
    \vspace{-0.3cm}
\end{figure}
\paragraph{Settings.} {According to the algorithms~\ref{algo:stage_1_2},~\ref{algo:stage_3}, \tool~consists of four key components: majority vote, jury selection, legal roles, and demonstrations. The jury selection algorithm ranks and selects high-quality LLMs; legal roles simulate human behavior by assigning responsibility to different models; and demonstrations provide in-context learning examples. We report the detection rates for various combinations on the Law-SG dataset.}

\noindent{\bf Results.} As illustrated in Figure~\ref{fig:ablation_study}, jury selection consistently improves detection rates, with particularly strong gains in Pool 1 and Pool 3. The impact of roles on detection rates is complex. In Pool 1 and Pool 2, combining roles with jury selection slightly reduces detection rates compared to using jury selection alone. However, in Pool 3, the inclusion of roles significantly boosts performance, resulting in the highest detection rate observed. In contrast, the use of demonstrations consistently enhances detection rates across all pools. Overall, combining all techniques does not always yield the highest detection rate, as seen in Pools 1 and 3 individually. However, it consistently ensures strong performance.

\section{Related Work}
\label{sect:related_work}
Our work is related to the literature on legal benchmarks, LLM vote, LLM debate, few-shot prompting, model editing, adversarial testing, and model alignment. The remaining related can be found in Appendix~\ref{sect:continue_related_work}.

\noindent{\bf Legal Benchmark.} LLMs are highly versatile and can be applied to unforeseen use cases, which makes their evaluation on legal domain particularly challenging. Lately, significant effort is being invested in this direction, with benchmarks being developed for various legal aspects of LLMs including EU AI act compliance (SELF-COMPL~\cite{guldimann2024compl}), expert-crafted legal tasks (LegalBench~\cite{guha2023legalbench}), Chinese legal evaluation (LexEval~\cite{li2024lexevalcomprehensivechineselegal}), European Court of Human Rights (ECtHR~\cite{chalkidis2019neural,chalkidis2021paragraph}), Supreme Court of the United States (SCOTUS~\cite{katz2017general}), European Union legislation (EUR-LEX~\cite{bowman2021will}), contract provision classification (LEDGAR~\cite{tuggener2020ledgar}), Unfair Terms of Service (UNFAIR-ToS~\cite{akash2024unfair}), and Case Holdings on Legal Decisions (CaseHOLD~\cite{zheng2021does}). While beneficial for LLM research in legal domain, these works lack of automated interpretation of local regulations and do not provide exhaustive coverage across all relevant aspects.

\noindent{\bf LLM Vote.} While LLMs provide useful answers to direct queries, they have been found to be more reliable and accurate when their answers are aggregated from multiple LLM responses. LLM vote involves aggregating answers from several LLMs to achieve a consensus. Majority vote is a multi-path voting method~\cite{wang2022self, li2022making, xue2023dynamic, wang2024soft, chen2023universal} that enhances the reliability and accuracy of the final outcome. Self-Consistency~\cite{wang2022self} involves sampling multiple answer paths to a single question and aggregating them to produce a final, more reliable answer. DiVeRSe~\cite{li2022making} is powered by a verifier to evaluate each reasoning step to filter out bad answers from a pool of reasoning paths. Dynamic Voting~\cite{xue2023dynamic} applies early exiting for problems that LLMs can confidently solve. Universal Self-Consistency~\cite{chen2023universal} uses LLMs to select the most consistent answer among multiple reasoning paths.
\section{Limitations}
Our main limitation lies in the dataset creation process. Real-world case law often involves multiple regulations and contains nuanced details that are difficult to assess, even for legal experts. Without additional context or clearer evidence, it can be challenging to identify violations accurately. In contrast, our data generation approach simplifies the problem by focusing on a single, clearly defined misconduct. Judgments are primarily made by LLMs following precise and structured instructions.
\section{Conclusion}
In this work, we explored various strategies for data generation and violation detection. Our results demonstrate that adversarial data effectively reveals differences in LLM performance, particularly in identifying violations. Furthermore, \tool, our jury-based voting method, consistently outperforms traditional majority voting by enhancing both detection accuracy and answer reliability.
\bibliographystyle{abbrv}
\bibliography{refs}
\newpage
\appendix
\section{Experimental Details}
\subsection{Jury Pool}
\begin{table}[h!]
    \centering
    \begin{tabular}{llllll}
    \toprule
    \multicolumn{2}{c}{$P_1$} & \multicolumn{2}{c}{$P_2$} & 
    \multicolumn{2}{c}{$P_3$} \\
    \cmidrule(l){1-2}
    \cmidrule(l){3-4}
    \cmidrule(l){5-6}
    Role & LLM & Role & LLM & Role & LLM \\
    \cmidrule(l){1-2}
    \cmidrule(l){3-4}
    \cmidrule(l){5-6}
    Lawyer & {\it Phi4} & Judge & {\it Qwen2.5} & Judge & {\it Llama3.1} \\
    Judge & {\it Mistral$^*$} & Judge & {\it Mistral$^*$} & Judge & {\it Mistral}\\
    Judge & {\it Llama3.1$^*$} & Lawyer & {\it Mistral$^*$} & Judge & {\it Phi4}\\
    Prosecutor & {\it Llama3.1$^*$} & Judge & {\it Mistral} & Prosecutor & {\it Llama3.1}\\
    Prosecutor & {\it Llama3.1} & Prosecutor & {\it Phi4} & Lawyer & {\it Mistral}\\
    Lawyer & {\it Mistral} & Judge & {\it Llama3.1$^*$} & Lawyer & {\it Phi4}\\
    \midrule
    \multicolumn{2}{c}{$P_4$} & \multicolumn{2}{c}{$P_5$} & 
    \multicolumn{2}{c}{$P_6$} \\
    \cmidrule(l){1-2}
    \cmidrule(l){3-4}
    \cmidrule(l){5-6}
    Role & LLM & Role & LLM & Role & LLM \\
    \cmidrule(l){1-2}
    \cmidrule(l){3-4}
    \cmidrule(l){5-6}
    Prosecutor & {\it Qwen2.5$^*$} & Lawyer & {\it Mistral$^*$} & Lawyer & {\it Llama3.1} \\
    Prosecutor & {\it Mistral$^*$} & Prosecutor & {\it Llama3.1$^*$} & Judge & {\it Qwen2.5} \\
    Prosecutor & {\it Llama3.1} & Prosecutor & {\it Qwen2.5$^*$} & Lawyer & {\it Qwen2.5$^*$} \\
    Judge & {\it Qwen2.5$^*$} & Judge & {\it Llama3.1} & Prosecutor & {\it Mistral$^*$} \\
    Lawyer & {\it Phi4} & Prosecutor & {\it Llama3.1} & Judge & {\it Qwen2.5$^*$} \\
    Prosecutor & {\it Mistral} & Lawyer & {\it Phi4} & Prosecutor & {\it Llama3.1$^*$} \\
    \midrule
    \multicolumn{2}{c}{$P_7$} & \multicolumn{2}{c}{$P_8$} & 
    \multicolumn{2}{c}{$P_9$} \\
    \cmidrule(l){1-2}
    \cmidrule(l){3-4}
    \cmidrule(l){5-6}
    Role & LLM & Role & LLM & Role & LLM \\
    \cmidrule(l){1-2}
    \cmidrule(l){3-4}
    \cmidrule(l){5-6}
    Lawyer & {\it Mistral$^*$} & Prosecutor & {\it Phi4} & Judge & {\it Phi4} \\
    Prosecutor & {\it Mistral$^*$} & Judge & {\it Qwen2.5$^*$} & Lawyer & {\it Mistral} \\
    Judge & {\it Qwen2.5} & Judge & {\it Qwen2.5} & Lawyer & {\it Mistral$^*$} \\
    Lawyer & {\it Phi4} & Judge & {\it Mistral$^*$} & Lawyer & {\it Qwen2.5$^*$} \\
    Prosecutor & {\it Mistral} & Lawyer & {\it Llama3.1$^*$} & Judge & {\it Qwen2.5} \\
    Judge & {\it Phi4} & Prosecutor & {\it Llama3.1} & Lawyer & {\it Llama3.1$^*$} \\
    \bottomrule
    \end{tabular}
    \vspace{0.3cm}
    \caption{Jury pools: models marked with * are fine-tuned versions.}
    \label{tab:jury_pools}
\end{table}

Table~\ref{tab:jury_pools} presents the jury pool configurations for three datasets: {\it Law-SG} ($P_1$--$P_3$), {\it Case-SG} ($P_7$--$P_9$), and Unfair-TOS~\cite{akash2024unfair} ($P_3$--$P_5$). Each pool consists of six members, with roles (Lawyer, Prosecutor, or Judge) randomly assigned to each member.

\subsection{Datasets}
\label{sect:datasets}
\begin{table}[h!]
    \centering
    \begin{tabular}{cp{8cm}cc}
    \toprule
    \# & Name & \# Pages & \# Misconducts \\
    \midrule
    1 & The Liquor Control (Supply and Consumption) (Liquor Licensing) Regulations &  12 & 8 \\
    2 & Parking Places Rules Regulations & 21 & 22 \\
    3 & Regulation of Imports and Exports (Chewing Gum) Regulations & 6 & 5 \\
    4 & Rapid Transit Systems Regulations & 24 & 31 \\
    5 & Smoking (Prohibition in Certain Places) Regulations & 27 & 5 \\
    \bottomrule
    \end{tabular}
    \vspace{0.3cm}
    \caption{Five Singapore Regulations}
    \label{tab:singapore_regulation_list}
\end{table}

\begin{table}[t]
    \centering
    \small
    \begin{tabular}{p{1.5cm}p{1.7cm}p{5cm}p{4cm}}
    \toprule
     Dataset & \# Examples & Scenario & Misconduct  \\
     \midrule
     {\it Law-SG}  & 3150  & While shopping at a mall, a driver parks their car in a designated season parking place, but they are unable to display their valid season parking ticket as their wallet was stolen during their visit. & Parking in a season parking place without possession of a season parking ticket \\ \midrule
     {\it Unfair-TOS} &  1137 & Your use of the services following that date constitutes your acceptance of the terms and conditions of the terms as modified. & This stipulates how the customer is legally obligated to adhere to the terms and conditions of utilizing a certain service that is provided by the service provider, regardless of necessitating explicit acknowledgment or acceptance of the conditions of use. All the clauses that gave service providers such rights were marked as unfair (1). \\ \midrule
     {\it Case-SG} & 58 & In March 2024, a man was seen vaping openly on an MRT train while leaning against the doors. He nearly fell when the doors opened at Braddell Station. & Prohibition of Tobacco Products Act (PTA): Using e-vaporisers in public places is illegal in Singapore with a fine of up to S\$2,000.\\ \midrule
     {\it DROP } & 630 & To start the season, the Lions traveled south to Tampa, Florida to take on the Tampa Bay Buccaneers. The Lions scored first in the first quarter with a 23-yard field goal by Jason Hanson. The Buccaneers tied it up with a 38-yard field goal by Connor Barth, then took the lead when Aqib Talib intercepted a pass from Matthew Stafford and ran it in 28 yards. The Lions responded with a 28-yard field goal. In the second quarter, Detroit took the lead with a 36-yard touchdown catch by Calvin Johnson, and later added more points when Tony Scheffler caught an 11-yard TD pass. Tampa Bay responded with a 31-yard field goal just before halftime. The second half was relatively quiet, with each team only scoring one touchdown. First, Detroit's Calvin Johnson caught a 1-yard pass in the third quarter. The game's final points came when Mike Williams of Tampa Bay caught a 5-yard pass. The Lions won their regular season opener for the first time since 2007 & N/A \\
     \bottomrule
    \end{tabular}
    \vspace{0.3cm}
    \caption{Dataset Description.}
    \label{tab:dataset_description}
\end{table}

\noindent {\bf Law-SG.} The dataset is constructed by generating 1,000 initial scenarios, each refined through five iterations of adversarial testing, resulting in a total of 3,150 scenarios (Table~\ref{tab:dataset_description}). Each scenario is the outcome of targeted adversarial attacks against four mainstream LLMs (Llama, Phi, Mistral, and Qwen) using misconducts drawn from five Singapore regulations (Table~\ref{tab:singapore_regulation_list}).
\bigbreak

\noindent {\bf Unfair-TOS.} The dataset consists of 50 Terms of Service (ToS) documents collected from major online platforms such as YouTube, eBay, and Facebook. The dataset is annotated at the sentence level with eight distinct categories of potentially unfair contractual terms, clauses that may infringe upon user rights as defined by EU consumer protection law.
\bigbreak

\noindent {\bf Case-SG}. The dataset comprises 58 real-world scenarios carefully curated from official Singapore newspapers. Each scenario reflects actual events or legal cases reported in the media, offering authentic context for evaluating legal reasoning and compliance with Singaporean law.
\bigbreak

\noindent {\bf DROP}. The DROP dataset (Discrete Reasoning Over Paragraphs) is a reading comprehension benchmark designed to evaluate a model's ability to perform discrete reasoning over text. It contains approximately 96,000 question-answer pairs sourced from Wikipedia, focusing on numerical reasoning, logical operations, and contextual understanding.
\bigbreak

\section{Implementation Details}
Our experiments were conducted on H100 GPU, each experiment will take several hours to days, depending on the number of inputs and the number of models. For example, it takes 8 hours to for 6 model votings on 630 examples.
\subsection{Prompt}
\begin{table}[h!]
    \centering
    \begin{tabular}{p{13cm}}
    \midrule
    Direct Scenario Generation Prompt \\
    \midrule
    Regulation: \{regulation\} \newline
    Misconduct: \{misconduct\} \newline
    Generate a real-life scenario that violates the misconduct. \newline
    The generated scenario should not exceed 5 sentences. \newline
    Do not include any explanations, only return a scenario. \\
    \midrule
    Chain-of-Thought Violation Detection Prompt \\
    \midrule
    Scenario: \{scenario\} \newline
    Question: Does it involve a violation of law? \newline
    The scenario happens in Singapore. \newline
    Return the answer (Yes/No) at the end of the response after a separator \#\#\#\#. \newline
    For example, \#\#\#\# Answer: Yes \newline
    Think step by step to answer the question. \\
    \midrule
    Jury Ranking Prompt \\
    \midrule
    \{context\} \newline
    --- \newline
    Scenario: \{scenario\} \newline
    Misconduct: \{regulation\} > \{misconduct\} \newline
    Instruction: \newline
    The scenario happens in Singapore. \newline
    It involves a violation of the specified misconduct.\newline
    Evaluate the correctness of \{label\} on a scale from 0.0 to 1.0. \newline
    Assign a unique, non-repeating score to each answer. \newline
    Return the answer (array of tuple) at the end of the response after a separator \#\#\#\#. \newline
    For example, \#\#\#\# [(0, 0.5), (1, 0.2), (2, 0.0)] \\
    \midrule
    Jury Vote Prompt \\
    \midrule
    Example: \newline
    Misconduct: \{misconduct\} \newline
    Scenario: \{scenario\_\} \newline
    Question: Does it involve a violation of law? \newline
    Answer: Yes \newline
    --- \newline
    Scenario: \{scenario\}\newline
    Question: Does it involve a violation of law?\newline
    The scenario happens in Singapore. \newline
    Return the answer (Yes/No) at the end of the response after a separator \#\#\#\#.\newline
    For example, \#\#\#\# Answer: Yes\newline
    Think step by step to answer the question. \\
    \bottomrule
    \end{tabular}
    \vspace{0.3cm}
    \caption{List of used prompts}
    \label{tab:list_of_used_prompts}
\end{table}

\subsection{Model Editing Details}
\label{sect:model_editing_details}

The model editing task~\cite{wang2024wise} involves making numerous updates to a pre-trained model over time, ensuring that it consistently refreshes its knowledge and stays aligned with the fast-changing information encountered in everyday life. This task modifies an initial base model $f_{\theta_0}$, parameterized by $\theta$ at the time step $0$, using a dataset $D_{\textrm{edit}} = \{(\mathcal{X}_e, \mathcal{Y}_e)~|~(x_1, y_1), \cdots,(x_T, y_T) \}$. Formally, at the time step $T$, the model editor, denoted by $\textrm{ME}$, inserts the T-th edit into the model $f_{\theta_{T-1}}$ and produces an edited model $f_{\theta_T}$. Let $\mathcal{P}(\cdot)$ be a function that rephrases $x$ to a set of semantic equivalent inputs (we assume $x \in \mathcal{P}(x)$). The task of model editing is defined as follows:
\begin{equation}
\label{eq:model_editing}
f_{\theta_T} = \textrm{ME}(f_{\theta_{T-1}}, x_T, y_T) ~ \textrm{s.t.} ~ f_{\theta_T}(x) = \begin{cases}
y_e & \textrm{ if } x \in \mathcal{P}(x_e) \wedge (x_e,y_e) \in {D}_{edit}\\
f_{\theta_0}(x) & \textrm{ otherwise}.\\
\end{cases}
\end{equation}

The edited model $f_{\theta_T}$ should produce a desired output $y_e$ for each in-scope input $x \in \mathcal{P}(x_e)$ and $(x_e,y_e) \in {D}_{edit}$, while maintaining the original model's performance $f_{\theta_0}(x)$ on an irrelevant input $(x,y) \in D_{irr}$ where $D_{irr}=\{(x, y) \mid x \notin \mathcal{P}(x_e), \forall x_e \in \mathcal{X}_e\}$. It also preserves knowledge from past edits $(x_{<T}, y_{<T}) \in D_{\textrm{edit}}$. Additionally, the result of applying $f_{\theta_{T}}$ to $x$ and $\mathcal{P}(x)$ should be identical. 

To measure the efficiency of a model editor, the edited model is subject to evaluation using the following metrics.

\bigbreak
\noindent\textbf{Reliability}: The edited model $f_{\theta_T}$ should generate the expected responses on intended edits:\\
\vspace{-3mm}
\begin{align}
\mathbb{E}_{(x_e,y_e) \in D_{\textrm{edit}}}~~\mathbbm{1}\{\textrm{argmax}_y f_{\theta_T}(y \mid x_e) = y_e\} 
\end{align}

\noindent\textbf{Locality}: The edited model $f_{\theta_T}$ should retain original responses on inputs that are irrelevant to intended edits:\\
\vspace{-3mm}
\begin{align}
\mathbb{E}_{(x, y) \in D_{\textrm{irr}}}~~\mathbbm{1}\{\textrm{argmax}_y f_{\theta_T}(y \mid x) = f_{\theta_0}(y \mid x)\} 
\end{align}


\noindent\textbf{Generality}: The model $f_{\theta_T}$ should generalize edits over other semantic equivalent inputs:\\
\vspace{-3mm}
\begin{align}
\mathbb{E}_{(x_e,y_e) \in D_{\textrm{edit}}}~~\mathbbm{1}\{\textrm{argmax}_y f_{\theta_T}(y \mid x) = y_e\}~\textrm{s.t.}~x \neq x_e \land x \in \mathcal{P}(x_e)
\end{align}

Our implementation is built on top of EasyEdit~\cite{jiang2024easyedit}, one of the most widely used frameworks for model editing. In our approach, the editing sentence is $p \oplus x_i \oplus y_i$ where $x_i$ is the scenario and $y_i$ is the expected answer. For example, $x_i$ is {\it ``Jason eats a burger on train''} and $y_i$ is {\it ``Answer: Yes. Reason: eating on trains is prohibited in Singapore''}. 

\subsection{Alignment Details}
\label{sect:alignment_details}
Since the original test split exclusively requires ``Yes'' responses, we augment the dataset with an equal number of ``No'' instances sampled from the DROP benchmark, ensuring a balanced distribution of affirmative and negative examples for robust evaluation. This approach mitigates potential bias and strengthens the reliability of our findings.
It is noted that each training input is a triplet $\langle x_i,y_w, y_l \rangle$ where $y_w$ and $y_l$ denote the preferred and dispreferred outputs. For example, $x_i$ is {\it ``Jason eats a burger on train''}, $y_l$ is {\it ``Answer: No. Reason: eating on trains is not a violation.''} and $y_w$ is {\it ``Answer: Yes. Reason: eating on trains is prohibited in Singapore''}. Our implementation utilizes Unsloth~\cite{unsloth} to perform Direct Preference Optimization (DPO)~\cite{rafailov2023direct} on a LoRA module, which is then integrated into the original model for inference. The DPO configuration is batch\_size=4, max\_steps = 20, num\_train\_epochs = 3, rank = 16, beta = 0.3.

\subsection{RAG Details}
\label{sect:rag_details}

We have implemented two distinct approaches for handling queries related to regulation documents. The first approach involves querying the raw text of regulation documents to extract relevant misconduct that can answer the question based on the provided scenario. However, this method tends to have lower accuracy, primarily due to incomplete or irrelevant text being retrieved, such as non-violative definitions or unrelated sections that do not address the specific query at hand. To mitigate these limitations, we have opted for the second approach, where we provide all collected misconduct as context. By using Langchain~\cite{langchain} for query processing, we can ensure more accurate and contextually relevant results, improving the overall performance of the system.

\section{Related Work}
\label{sect:continue_related_work}
\paragraph{LLM Debate.}{LLM agents are AI agents that harness the reasoning ability of LLMs to complete tasks automatically. The idea of LLM debate~\cite{du2023improving, irving2018ai,khan2024debating, michael2023debate, wu2023autogen} is to focus on exploring various strategies for facilitating conversations between two or more LLM agents to enhance the final outcomes. The paper~\cite{du2023improving} examines a multi-round debate between multiple LLMs, demonstrating enhanced reasoning and factual accuracy. The AI safety work~\cite{irving2018ai} proposes a game of LLM debate for training a safe AI system. The paper~\cite{khan2024debating} presents a framework where both experts and non-experts engage in debate over questions, with non-experts ultimately selecting the best answer. The results demonstrate that such debates enhance the accuracy of non-expert models. The work~\cite{michael2023debate} demonstrates that debate between two unreliable experts can help a non-expert judge more reliably identify the truth. The emerging framework~\cite{wu2023autogen} enables researchers to rapidly prototype conversations between LLM agents with various speaking permission strategies such as round-robin, random, or manager-controlled turn-taking.}

\paragraph{Few-Shot Prompting.}{Complex reasoning via intermediate steps forms an essential aspect of LLMs. These steps can be elicited through instructions such as {\it ``Let's think step by step''}~\cite{kojima2022large} instruction or through manually curated demonstrations, where each consists of a question and a reasoning chain that leads to an answer. Recently, there has been substantial efforts~\cite{zhang2022automatic, lu2023chain,wang2023boosting,wang2023cue,press2022measuring,kim2022self,su2022selective} in selecting or synthesizing high-quality demonstrations. Auto-CoT~\cite{zhang2022automatic} samples questions with diversity and generates reasoning chains to construct demonstrations. It outperforms manual chain-of-thought on eight out of ten datasets. CoD~\cite{lu2023chain} enhances language models' translation capabilities by augmenting prompts with word-meaning examples (e.g., ``haben'' means ``avoir''). This approach successfully improves the translation performance of both ChatGPT and InstructGPT. CoK~\cite{wang2023boosting} uses prompts with two main components: {\it evidence triples} and {\it explanation hints}. Evidence triples reflect the overall reasoning path from query to answer, while explanation hints provide explanations for this evidence. CoK evaluates the reliability of reasoning chains based on {\it factuality} and {\it faithfulness}, and rethinks those deemed unreliable. Cue-CoT~\cite{wang2023cue} extracts linguistic cues from dialogue context through intermediate reasoning, then uses these cues to craft more tailored responses. Vote-K~\cite{su2022selective} is a graph-based technique for selecting demonstrations. Rather than annotating the entire training dataset, a costly and time-consuming process. Vote-K efficiently identifies the most informative examples for annotation, which can then serve as demonstrations in few-shot prompting.}

\paragraph{Model Editing.}{Regular updates are essential to keep LLMs aligned with evolving knowledge. Model editing methods ~\cite{meng2023memit, gupta2024unified, Meng2022LocatingAE, wang2024wise, hartvigsen2023aging,li2023pmet, yu2024melo, mitchell2022memory, nguyen2024uniadapt, wang2024memoe, wang2024lemoe} enable LLMs to incorporate the latest knowledge in a timely and effective manner. ROME~\cite{Meng2022LocatingAE}, EMMET~\cite{gupta2024unified}, PMET~\cite{li2023pmet} identify layers in LLMs responsible for recalling factual knowledge, then edit their feedforward weights to update the identified factual associations. GRACE~\cite{hartvigsen2023aging}, MELO~\cite{yu2024melo}, and SERAC~\cite{mitchell2022memory} store updated knowledge in memory and retrieve it using similarity-based queries when requested. WISE~\cite{wang2024wise}, UniAdapt~\cite{nguyen2024uniadapt}, MEMoE~\cite{wang2024memoe}, and LEMoE~\cite{wang2024lemoe} utilize a Mixture of Experts (MoE) architecture to route inputs to specific experts, enabling efficient retrieval and updating of new knowledge. Despite highly efficient reported results, recent studies~\cite{gu2024model, gupta2024model, li2024should} highlight significant weaknesses in model editing. These methods often degrade the general capabilities of large language models (LLMs) and lead to forgetting previous edits as the number of edits reaches thousands. Furthermore, no proposed approaches have addressed editing large-scale models, such as those with 70B parameters or more.}

\paragraph{Adversarial Testing.}{This line of research~\cite{zheng2024ali, perez2022red, yu2023gptfuzzer, sun2024principle, mehrotra2023tree} explores the use of LLMs to generate data for evaluating or attacking other LLMs. LLM-based red teaming~\cite{perez2022red} generates test questions intended to elicit misaligned responses from target models, such as offensive language, data leakage, or distributional bias. GPT-FUZZER~\cite{yu2023gptfuzzer} is a template-based fuzzing framework that employs LLMs to mutate test inputs and assess the robustness of other LLMs. SELF-ALIGN~\cite{sun2024principle} utilizes LLMs to synthesize alignment data with minimal human supervision. TAP~\cite{mehrotra2023tree} employs a red-team LLM to iteratively refine candidate attack prompts until a successful jailbreak is achieved. ALIGN~\cite{sun2024principle} focuses on generating long-tailed scenarios in which violations are subtly embedded, aiming to deceive the target LLM through a refinement loop powered by a strong generative model.}

\paragraph{Model Alignment.}{Pretrained LLMs capture general language patterns, structures, and knowledge from extensive datasets, but require further refinement to align with human preferences. To address misalignment, advanced alignment techniques~\cite{ouyang2022training, rafailov2024direct, hong2024orpo, xu2024contrastive, shao2024deepseekmath} have been developed to mitigate the misaligned responses. RLHF~\cite{ouyang2022training} uses a reward model to optimize the LLM's behavior. DPO~\cite{rafailov2024direct} learns human preferences from pairs of preferred and dispreferred answers, eliminating the need for reward models or reinforcement learning. ORPO~\cite{hong2024orpo} integrates supervised fine-tuning (SFT) and preference optimization into a single phase, eliminating the need for a reference model or separate alignment stages. GRPO~\cite{shao2024deepseekmath} enhances group-level robustness in preference optimization by adapting LLMs to diverse group preferences. Unlike DPO, which targets average performance, GRPO optimizes for worst-case group performance, ensuring equitable outcomes across groups. Bolstered by promising outcomes in reasoning models, alignment methods are garnering growing interest from the research community.}

\end{document}